\documentclass[mlabstract,final]{jmlr}

\usepackage{booktabs}
\usepackage{graphicx}
\setkeys{Gin}{draft=false}
\usepackage[load-configurations=version-1]{siunitx}
\usepackage{tikz}
\usepackage{placeins}
\usepackage{float}
\usetikzlibrary{arrows.meta,calc,angles,quotes,positioning}



\jmlrvolume{}
\firstpageno{1}
\editors{}
\jmlrproceedings{NeurReps}{NeurReps -- Extended Abstract Track}
\jmlryear{2026}
\jmlrworkshop{Symmetry and Geometry in Neural Representations}

\title[Local Action Geometry in LeWM]
{Does LeWorldModel Preserve Local Action Geometry?}

\author{\Name{Vasil Vasilev} \Email{vasil.g.vasilev.2022@elsys-bg.org}\\
\addr Technological School Electronic Systems (TUES)
\AND
\Name{Siddharth Nayak} \Email{sidnayak@mit.edu}\\
\addr Massachusetts Institute of Technology
}

\hypersetup{
  pdftitle={Does LeWorldModel Preserve Local Action Geometry?},
  pdfauthor={Vasil Vasilev; Siddharth Nayak},
  pdfsubject={Symmetry and Geometry in Neural Representations, 2026},
  pdfkeywords={world models, JEPA, action sensitivity, representation geometry, Gram matrix}
}

\begin{document}

\maketitle

\begin{abstract}
We examine how closely pretrained LeWorldModel (LeWM) reproduces the
effects of small action changes. Using matched simulator and model
responses, we measure action sensitivity with Gram matrices at three
perturbation scales. Across nearly a thousand states from Reacher and
OGBench Cube, the most sensitive action directions differ by median
angles of about $36^\circ$ and $42^\circ$, respectively, at the smallest
scale. Increasing the perturbation tenfold reduces median normalized
Gram discrepancy by roughly 70\% on Reacher and 47\% on Cube.
Geometry measured in pixels and embeddings varies more across scales
than the chosen physical-state geometry. On held-out directions,
environment Grams achieve median embedding response errors of 17\% and
28\%, compared with 26\% and 38\% for an isotropic reference with the same
trace. LeWM's errors are higher at this scale. These findings support
scale-specific evaluation of action sensitivity; its effect on control
remains an open question.
\end{abstract}


\section{Introduction}
\label{sec:intro}
World models support planning through learned predictive dynamics
\citep{bib1,bib2,bib3}. Joint-Embedding Predictive Architectures learn in
representation space \citep{bib4,bib5}, including action-conditioned systems
such as V-JEPA~2 and LeWorldModel (LeWM) \citep{bib6,bib7}. Accurate prediction
at a sampled action does not establish that nearby action directions have
the correct relative effects. This motivates evaluating action sensitivity
alongside aggregate prediction and control scores.

We study pretrained LeWM on Reacher and OGBench Cube using simulator
counterfactuals and a shared frozen encoder. Related symmetry-aware and
geometric perspectives motivate this diagnostic \citep{bib8,bib9}.
Our contributions
are a matched finite-scale comparison, calibration against simple references,
and validation on held-out action directions in three distinct output
metrics. We distinguish sensitivity in the encoder's output space from
physical response and goal-directed action choice.

\section{Measuring Finite-Scale Action Geometry}
\label{sec:method}
Let $H_t$ be a fixed three-observation context, $E_\theta$ the frozen encoder,
and $A_t=(a_t,\ldots,a_{t+4})$ the recorded five-action block. We perturb all
five actions by a shared normalized offset $u\in\mathbb R^d$:
\[
 A_t(u)=(a_t+Du,\ldots,a_{t+4}+Du),\qquad
 D=\operatorname{diag}(\text{action standard deviations}).
\]
$F_t^{\rm model}(u)$ predicts the next embedding from $H_t$ and $A_t(u)$;
$F_t^{\rm env}(u)$ encodes the rendered endpoint after executing the same
block from the restored simulator state. Both maps use the same output
space. With $e_i$ the $i$th coordinate vector, define
\[
\begin{aligned}
 J_{t,\epsilon}^{b}e_i=
 &\frac{F_t^b(h_i e_i)-F_t^b(-h_i e_i)}{2h_i},
 \qquad h_i=\frac{\epsilon(a_i^{\max}-a_i^{\min})}{2D_{ii}},\\
 G_{t,\epsilon}^{b}={}&(J_{t,\epsilon}^{b})^\top J_{t,\epsilon}^{b}.
\end{aligned}
\]
where $b\in\{\mathrm{env},\mathrm{model}\}$. The shared offset gives $d$
action coordinates. The identity
$\|J_{t,\epsilon}^{b}u\|^2=u^\top G_{t,\epsilon}^{b}u$
motivates predicting response magnitudes. We test this approximation at
each scale because contacts and rasterization can limit local linearity.

We measure relative sensitivity using
$\widetilde G=G/\operatorname{tr}(G)$ and
$D_G=\|\widetilde G^{\rm model}-\widetilde G^{\rm env}\|_F$.
Zero or nonfinite traces are invalid; the primary additive constant is zero.
We separately report absolute gain, condition numbers and eigenspace angles.
Calibration compares $D_G$ with isotropy ($I_d/d$), random orthogonal
conjugation of the environment metric, and shuffled-state matching.
Each reference removes a specific source of information.
Definitions and episode-bootstrap uncertainty are in
Appendix~\ref{apd:metrics}.

One released checkpoint per environment is evaluated without training.
The common no-clipping bank retains 458 Reacher anchors ($d=2$; 42
action-bound exclusions) and 500 Cube anchors ($d=5$). Five wrapper steps
span 0.20 and 0.25\,s, respectively. We report $\epsilon=10^{-3}$ (primary),
$3\times10^{-3}$ and $10^{-2}$. Following an audit of action alignment,
output coordinates and state restoration, we regenerate trajectories in
the current simulator from the original dataset sampling anchors.
The revised estimates supersede the earlier results
(Appendix~\ref{sec:robustness}). Appendix~\ref{apd:response-replay}
documents replay exceptions and the resulting protocol change.

\section{Results}
\label{sec:results}
\paragraph{The model captures some action structure but remains misaligned.}
The comparison separates sensitivity strength from action-direction
alignment. Eigenvalues describe the squared response strengths along
the corresponding eigenvectors; similar condition numbers can therefore
coexist with different sensitive directions. Trace-normalized $D_G$
captures both relative spectral shape and orientation, while deliberately
discarding a common multiplicative gain.

At the primary scale, the most sensitive directions differ by a median
$36^\circ$ on Reacher and $42^\circ$ on Cube. The latter also shows about
$40^\circ$ of misalignment between its leading two-dimensional subspaces
(Figure~\ref{fig:local-metric-ellipses}). Median $D_G$ is 0.465 on Reacher
and 0.508 on Cube; their 95\% episode-bootstrap intervals are
$[0.423,0.500]$ and $[0.487,0.529]$.

Reacher's environment and model condition numbers are both near 4.5,
despite their directional mismatch. Cube's median model condition number
is roughly four times the environment value. The numerical floor affects
about a third of Cube environment matrices, so the upper tail requires
care. Substantial misalignment remains when we restrict the comparison
to well-separated leading eigenvalues (Appendix~\ref{sec:robustness}).

\begin{table}[t]
\floatconts{tab:calibration}
{\caption{Median primary-scale normalized Gram discrepancy (lower is closer).
Randomized references use 1,000 draws per state.}}
{\small\begin{tabular}{@{}lrrrr@{}}
\toprule
Environment & Model & Isotropic & Rotation & Shuffle\\
\midrule
Reacher & 0.465 & 0.449 & 0.632 & 0.530\\
Cube & 0.508 & 0.609 & 0.885 & 0.560\\
\bottomrule
\end{tabular}}
\end{table}

Cube has lower discrepancy than all three references. Reacher improves
on rotation and shuffle, while isotropy is closer to its environment metric
(Table~\ref{tab:calibration}).
Paired model-minus-shuffle median differences are $-0.059$ for Cube
(95\% interval $[-0.074,-0.029]$) and $-0.047$ for Reacher
($[-0.071,-0.014]$). Thus matching the state helps within this bank.
Absolute sensitivity is also lower in the model. The median ratio
$\sqrt{\operatorname{tr}(G^{\rm model})/\operatorname{tr}(G^{\rm env})}$
indicates about half the environment's sensitivity on Cube and four-fifths
on Reacher.

\begin{figure}[t]
\vspace{-2pt}
\floatconts
  {fig:local-metric-ellipses}
  {\caption{Action-sensitivity ellipses at $\epsilon=10^{-3}$.
  We select the sampled state nearest the median $D_G$ in each environment. Each curve maps the unit circle
  through $B^{1/2}$, where $B$ is the trace-normalized Gram restricted to
  the displayed plane. Reacher shows both action dimensions. Cube shows
  the environment's leading two-dimensional plane, which cannot reveal
  all misalignment in its five-dimensional action space.}}
  {\centering
  \includegraphics[draft=false,width=0.77\linewidth]
  {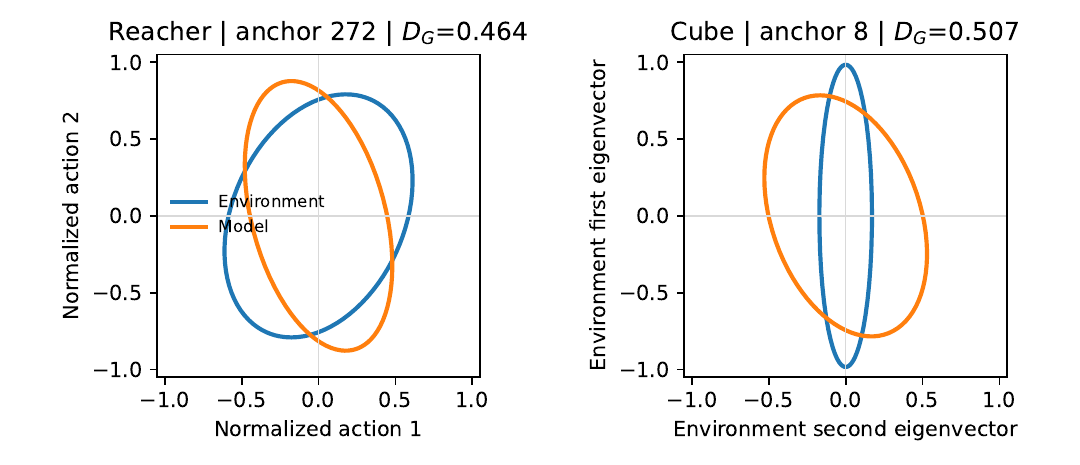}}
\end{figure}

\paragraph{Visual geometry changes more with perturbation scale.}
A tenfold increase in perturbation size reduces median $D_G$ by about
70\% on Reacher and 47\% on Cube. To investigate this scale dependence,
we measure the same simulator responses in scaled physical pose, RGB
pixels and frozen embeddings. Reacher's median physical-pose Gram
change is below one millionth, while its pixel and embedding changes
are about 0.11 and 0.43. Cube shows the same ordering: approximately
0.03, 0.08 and 0.49, respectively.

These comparisons show how strongly the output metric affects the
measurement. Rendering, encoding and nonlinear dynamics may each
contribute; their individual effects remain unresolved. Physical-response
outliers are retained (Appendix~\ref{apd:response}). We report all three
scales because changing the perturbation changes the response under study.

\paragraph{The Gram predicts response sizes along held-out directions.}
For eight held-out sphere directions per state, reused across scales, we
compare $\hat r(u)=\sqrt{u^\top G_\epsilon u}$ with the centered response
$r(u)=\|F(u)-F(-u)\|/2$. Coordinate Grams use none of these outcomes.
We measure weighted absolute percentage error (WAPE) as
$\sum_j|\hat r_j-r_j|/\sum_jr_j$ for each state, then report the median
across states. At the primary scale, environment embedding Grams have
lower error than isotropy, while model Grams have higher error
(Table~\ref{tab:response}). The 95\% intervals for the paired median
advantage of environment Grams span roughly six to eight percentage
points on Reacher and eight to twelve on Cube.

\begin{table}[t]
\floatconts{tab:response}
{\caption{Held-out embedding response WAPE (\%, lower is better),
$\epsilon=10^{-3}$. Isotropy uses the environment trace;
oracle gain rescales the model Gram to that trace.}}
{\small\begin{tabular}{@{}lrrrr@{}}
\toprule
Environment & Env.\ Gram & Isotropic & Model & Model + oracle gain\\
\midrule
Reacher & 17.0 & 26.2 & 33.2 & 26.6\\
Cube & 28.0 & 38.2 & 56.4 & 32.3\\
\bottomrule
\end{tabular}}
\end{table}

Gain error helps explain Cube's different outcomes on the two tests.
$D_G$ compares relative geometry, while WAPE also measures response size.
Giving the model the environment's trace reduces median WAPE by about
24 percentage points, bringing it below isotropy and within five points
of the environment Gram. This oracle correction uses simulator information,
so its role is to separate gain error from the remaining mismatch.

Physical-pose prediction has median WAPE below 0.04\% in both environments
at the primary scale. Pixel errors are about 5\% on Reacher and 15\% on
Cube. Predicting one-sided embedding displacements is harder than predicting
centered responses: median errors rise from 17\% to 32\% on Reacher and
from 28\% to 43\% on Cube. Appendix~\ref{apd:response} gives the full
results and retained failures.

\section{Discussion and Conclusion}
\label{sec:discussion}
LeWM captures part of the environment's action structure, with agreement
depending on perturbation scale and output metric. Held-out tests give
the Gram a concrete interpretation: it predicts response magnitudes
within the declared measurement space and scale. The original 50-state
subsets gave identical repeated results. Later image differences required
a documented change to the Cube replay protocol, which limits claims
about reproducibility. Additional checkpoints are needed to estimate
variation across training seeds.

The implications for training and control remain open.
For example, $J_{\rm model}=-J_{\rm env}$ gives equal Grams and opposite
outcome directions. Evaluating signed, goal-directed responses and
closed-loop performance would establish whether better sensitivity
estimates help action selection (Appendix~\ref{apd:isotropy-negative}).
\label{sec:main-end}
\clearpage
\bibliography{pmlr-sample}
\clearpage
\appendix

\section{Geometry Metrics and Additional Derivations}
\label{apd:metrics}

\subsection{Trace-normalized metric discrepancy}

Relative directional geometry is invariant to positive rescaling of $G_t$.
We therefore use
$\widetilde G_t=G_t/\operatorname{tr}(G_t)$ and
$D_G(t)=\|\widetilde G_t^{\mathrm{model}}-\widetilde G_t^{\mathrm{env}}\|_F$.
The primary trace-normalization constant is $\delta=0$.
We mark zero or non-finite traces as invalid. We separately test
additive constants $10^{-30},10^{-12},10^{-8}$. Small $D_G$ indicates similar relative sensitivity and orientation.
We assess absolute gain separately.
We also report
$S_b=\sqrt{\operatorname{tr}(G_t^b)}$ and $S_{\rm model}/S_{\rm env}$.
The shared frozen encoder gives this ratio a common latent scale.
Its interpretation depends on that representation.

\subsection{Eccentricity}

We define
$\kappa(G_t)=\lambda_{\max}(G_t)/\lambda_{\min}^{\mathrm{eff}}(G_t)$,
where
$\lambda_{\min}^{\mathrm{eff}}=
\max\{\lambda_{\min},10^{-6}\lambda_{\max},10^{-12}\}$.
We report the fraction of floor-limited matrices; condition numbers in that
subset should be read as capped estimates. For additional analysis we use
$D_\kappa(t)=\log\kappa(G_t^{\mathrm{model}})-\log\kappa(G_t^{\mathrm{env}})$
and $E_\kappa(t)=|D_\kappa(t)|$.
Positive $D_\kappa$ indicates eccentricity amplification relative to the
environment and negative values indicate eccentricity suppression.

\subsection{Directional and subspace alignment}

For dominant eigenvectors we use
$\theta_1(t)=\cos^{-1}(|(v_{t,1}^{\mathrm{env}})^\top v_{t,1}^{\mathrm{model}}|)$;
the absolute value removes the arbitrary sign of an eigenvector.

For a leading $k$-dimensional eigenspace, let
$V_{t,k}^{\mathrm{env}}=[v_{t,1}^{\mathrm{env}}\;\cdots\;v_{t,k}^{\mathrm{env}}]$
and define $V_{t,k}^{\mathrm{model}}$ analogously. If $\sigma_i$ are the
singular values of $(V_{t,k}^{\mathrm{env}})^\top V_{t,k}^{\mathrm{model}}$,
then the principal angles are $\theta_i=\cos^{-1}(\sigma_i)$. We use the
maximum principal angle for the leading two-dimensional Cube subspaces.
Near eigenvalue ties, these directions are not reliably identified. We report
the trace-normalized gaps $\widetilde\lambda_1-\widetilde\lambda_2$ and
$\widetilde\lambda_2-\widetilde\lambda_3$ (descending eigenvalues), and
condition angle summaries on both branches having gaps at least 0.05 or 0.1.
These checks describe angle sensitivity; all states remain in the $D_G$ analysis.

\subsection{Calibration and uncertainty}

We compare the model with three reference discrepancies:
\[
\begin{aligned}
 D_{\rm iso}(t)&=\|\widetilde G_t^{\rm env}-I_d/d\|_F,\\
 D_{\rm rot}(t,Q)&=\|\widetilde G_t^{\rm env}
                  -Q^\top\widetilde G_t^{\rm env}Q\|_F,\\
 D_{\rm shuffle}(t,\pi)&=\|\widetilde G_t^{\rm model}
                     -\widetilde G_{\pi(t)}^{\rm env}\|_F.
\end{aligned}
\]
Here $Q$ is Haar-distributed orthogonal and $\pi$ is a uniformly sampled
derangement of the retained states. Rotation preserves the environment's
spectrum while randomizing its action-coordinate orientation; shuffling
tests whether matching the current state helps relative to another sampled
state. Each reference removes a different source of information.
Isotropy provides a direction-agnostic comparison.
An isotropic matrix has no unique dominant direction, so its angle score
is undefined.

We use 1,000 draws, take each state's median null discrepancy, then summarize
across states. Paired differences subtract that per-state reference from
the model discrepancy. Confidence intervals use 2,000 episode-cluster
bootstrap resamples (seed 42). These intervals quantify uncertainty across
sampled anchors. Estimating training-seed uncertainty requires more checkpoints. The first 50 eligible anchors receive three
complete computations per scale. We report two-way single-measure
absolute-agreement ICC for repeats and separately across scales, together
with cross-scale Spearman rank correlations. Agreement across repeats
does not imply agreement across perturbation scales.

\subsection{Anchor drift in the new bank}

The coherent live state need not equal the archived step-10 state. For Cube,
the median per-state maximum coordinate difference is $5.88\times10^{-4}$
in qpos and $4.10\times10^{-4}$ in qvel; the respective maxima are 0.0262
and 2.2505. These audit values retain the coordinates' original units.
We keep all high-drift anchors. Reacher's maximum differences are below
$3\times10^{-15}$, indicating near-identical replay of its archived states.
Cube's drift means that the new bank samples somewhat different trajectories.

\subsection{Environment-relative interpretation}

A complementary theoretical comparison is
\[
R_t=
\left(G_t^{\mathrm{env}}+\eta I\right)^{-1/2}
G_t^{\mathrm{model}}
\left(G_t^{\mathrm{env}}+\eta I\right)^{-1/2},\qquad\eta>0.
\]
For positive-definite $G_t^{\mathrm{env}}$, taking $\eta=0$ and defining
$x=(G_t^{\mathrm{env}})^{1/2}\Delta a$ gives
$\Delta a^\top G_t^{\mathrm{env}}\Delta a=x^\top x$ and
$\Delta a^\top G_t^{\mathrm{model}}\Delta a=x^\top R_tx$.
Thus $R_t$ describes model geometry in coordinates where the environment
metric is locally Euclidean. In this unregularized positive-definite case, if
$G_t^{\mathrm{model}}=c_tG_t^{\mathrm{env}}$, then $R_t=c_tI$.
For Cube, $G_t^{\mathrm{env}}$ is often close to singular, so inverse-metric
quantities can amplify numerical noise; this is why $D_G$ and principal
subspace angles are our primary empirical comparisons.


\section{What Gram Agreement Does Not Establish}
\label{apd:isotropy-negative}

Gram agreement characterizes relative response magnitudes. Outcome
direction remains ambiguous: for example,
$J_{\rm model}=-J_{\rm env}$ gives identical Gram matrices but opposite latent
responses. More generally, left multiplication by an orthogonal output-space
map leaves $J^\top J$ unchanged. Trace normalization also removes overall
sensitivity scale. Goal-directed action preferences, planning efficiency and closed-loop
success therefore require separate measurements. Isotropy remains useful
as a comparison baseline. In an anisotropic environment, using it as a
training target would suppress genuine differences in action sensitivity.

\section{Bank Construction and Robustness}
\label{sec:robustness}

\begin{table}[ht]
\vspace{-8pt}
\floatconts
  {tab:env-model}
  {\caption{New-bank finite-scale geometry at $\epsilon=10^{-3}$.
Values are median [IQR] where reported; $N=458$ Reacher, $N=500$ Cube.
Lower $D_G$ and angles indicate closer agreement; $\kappa$ describes
anisotropy.}}
  {\scriptsize
  \setlength{\tabcolsep}{2.7pt}
  \renewcommand{\arraystretch}{0.82}
  \begin{tabular}{@{}lcccccc@{}}
  \toprule
  Env. & $d_a$ & $D_G$ & $\kappa_{\rm env}$ & $\kappa_{\rm model}$
  & Top-1 & Top-2 \\
  \midrule
  Reacher & 2 & $0.465\,[0.270,0.708]$ & 4.49 & 4.50
  & $36.03\,[18.04,53.85]^\circ$ & -- \\
  Cube & 5 & $0.508\,[0.397,0.633]$ & $2.84{\times}10^3$
  & $1.07{\times}10^4$ & $41.52\,[25.03,66.09]^\circ$
  & $40.27\,[27.13,58.12]^\circ$ \\
  \bottomrule
  \end{tabular}}
\vspace{-8pt}
\end{table}

\paragraph{Matched transition construction.}
The current-simulator bank uses MuJoCo 3.11.0, OGBench 1.2.1 and
$224\times224$ images, with an opaque Cube arm and hidden target markers.
Each eligible anchor must reproduce its live context image exactly after
snapshot restoration, and two unperturbed branch rollouts must agree in
pixels exactly and in qpos/qvel within $10^{-12}$. The bank retains context
images, counterfactual images and embeddings, response matrices, anchor
identifiers, and checkpoint/source hashes. Historical results remain archived.
Full native simulator data and mutable wrapper state are restored before
each branch without an intervening forward call. These are the primary-bank
checks; the later response-validation replay protocol is separately
documented in Appendix~\ref{apd:response-replay}.

\paragraph{Cube finite-difference scale.}
Because rendering and contacts can introduce finite-scale effects, we repeat
both model and environment measurements over a ten-fold perturbation range.
Table~\ref{tab:cube-epsilon} reports Cube results; the cross-environment trends
appear in Figure~\ref{fig:cube-robustness}. Each scale defines its own
finite-response measurement; convergence to a smooth derivative remains untested.

\begin{table}[!ht]
\vspace{-2pt}
\floatconts
  {tab:cube-epsilon}
  {\caption{Cube robustness across finite-difference scales.}}
  {\scriptsize
  \setlength{\tabcolsep}{3.4pt}
  \renewcommand{\arraystretch}{0.86}
  \begin{tabular}{@{}cccc@{}}
  \toprule
  $\epsilon$ & Median $D_G$ & Top-1 angle & Top-2 max angle \\
  \midrule
  $10^{-3}$ & 0.508 & $41.52^\circ$ & $40.27^\circ$ \\
  $3{\times}10^{-3}$ & 0.408 & $34.82^\circ$ & $30.38^\circ$ \\
  $10^{-2}$ & 0.271 & $20.96^\circ$ & $18.48^\circ$ \\
  \bottomrule
  \end{tabular}}
\end{table}

\paragraph{Scope of the revision.}
The corrected Reacher measurements do not support the historical
direction-preserving/eccentricity-amplifying interpretation: both median
condition numbers are near 4.5, but the median dominant-direction angle
is $36.03^\circ$. Primary-scale paired model-minus-isotropic $D_G$
differences are $+0.083$ (95\% interval $[0.064,0.108]$) for Reacher and
$-0.108$ ($[-0.128,-0.094]$) for Cube. Paired median differences need not
equal differences of separately computed medians.

We do not transfer historical gripper-contact labels to the new trajectories.
The earlier contact/non-contact figure and claim are therefore not retained
as evidence for this bank. Resolving contact-specific effects requires
measuring contact events along these live counterfactuals.

\paragraph{Repeated runs agree, while changing scale changes the result.}
On Cube's first 50 eligible anchors, all three repeats produced identical
response matrices and $D_G$ at every scale (absolute-agreement ICC 1).
Across scales, however, $D_G$ has ICC 0.220 and pairwise Spearman correlations
0.220--0.432. Between $10^{-3}$ and $10^{-2}$, the environment's dominant
direction changes by a median $38.99^\circ$, versus $0.01^\circ$ for the model.
The model's leading direction stays nearly fixed while the environment
embedding's direction rotates by almost $39^\circ$. This supports reporting
each scale separately. Rendering, nonlinear dynamics and contact effects
remain possible contributors.
Reacher likewise repeats exactly on its 50-anchor subset, while its median
$D_G$ falls from 0.465 to 0.308 to 0.141 across the three scales; dominant
angles fall from $36.03^\circ$ to $20.70^\circ$ to $8.89^\circ$.
Its cross-scale ICC is 0.200 and rank correlations range from 0.205 to 0.375.

\paragraph{Spectral gaps and numerical floors.}
At the primary scale, requiring both Cube top-1 gaps to be at least 0.1
retains 449 states with median angle $41.09^\circ$; the corresponding top-2
gap check retains 302 with median maximum angle $34.66^\circ$.
The mismatch therefore is not explained solely by nearly tied eigenspaces.
The condition-number floor is active for 32.4\% of Cube environment metrics
and none of its model metrics, limiting interpretation of the upper tail.
Adding a trace constant up to $10^{-8}$ changes Cube $D_G$ by less than
$9\times10^{-10}$ at all three scales. Large condition numbers do not, by
themselves, imply a small trace or a dominant trace regularizer.
Neither Reacher branch is floor-limited at the primary scale; its maximum
trace-constant effect is below $3\times10^{-11}$ across scales. Requiring
both Reacher top-1 gaps to exceed 0.1 retains 454 states with median angle
$35.80^\circ$.

\begin{figure}[t]
\vspace{-2pt}
\floatconts
  {fig:cube-robustness}
  {\caption{Geometry measurements change with perturbation scale.
  Lines show medians; shading covers the middle half of states.
  Each environment uses the same eligible states at every scale.}}
  {\centering
  \includegraphics[draft=false,width=0.85\linewidth]
  {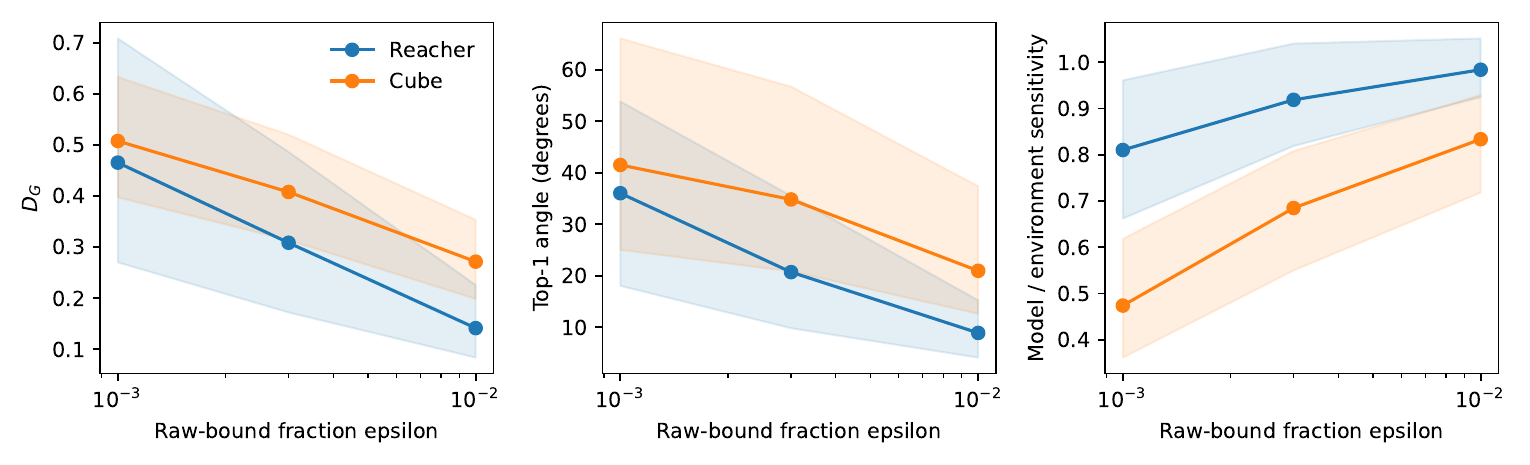}}
\end{figure}

\paragraph{Camera-ready protocol revision.}
These measurements supersede the historical estimates under a revised protocol. An audit found action-block/output-coordinate inconsistencies
in the available evaluation code and incomplete historical simulator-state
restoration. We therefore use the original datasets as \emph{sampling anchors}
for a new current-simulator bank: initialize at each anchor's context start,
execute ten recorded actions, and render fresh context frames at steps
0, 5 and 10. Counterfactuals branch from that live step-10 state, which can
differ from the archived state. This keeps context and outcomes internally
consistent within the current runtime. The original collection runtime
remains unavailable. Both branches now use matched finite differences;
the earlier description paired model autodiff with environment finite differences.

\section{Output-Space and Held-Out Response Validation}
\label{apd:response}

\subsection{Separate output metrics}
We repeat the coordinate and held-out perturbations on the same eligible
anchors and at all three declared scales. The physical-pose metric uses
MuJoCo's generalized-position tangent from the unperturbed endpoint to
each perturbed endpoint, computed with \texttt{mj\_differentiatePos} at
unit time. Translational components are divided by 0.1\,m and rotational
components by 1\,rad, then the vector is divided by $\sqrt{n_v}$.
Here $n_v=2$ for Reacher and $n_v=20$ for Cube, including robot, gripper
and object coordinates. This avoids treating quaternion signs as physical
displacements, but still makes an explicit, non-unique choice of physical
units and coordinate weights.

The supplemental pose-plus-velocity metric concatenates this vector with
$T(v-v_0)$ using the same component scaling and $1/\sqrt{n_v}$ factor,
then divides the concatenation by $\sqrt{2}$; $T$ is the five-step horizon.
Pixels use flattened RGB values divided by $255\sqrt{N_{\rm RGB}}$.
Embeddings use the frozen encoder's Euclidean metric without trace
normalization of output vectors. Grams are constructed independently in
each space. Gain depends on the chosen units and coordinate weights,
so comparisons across spaces require care.

Within each output metric, scale dependence is measured by
$\|\widetilde G_{t,0.001}^{\rm env}
-\widetilde G_{t,0.01}^{\rm env}\|_F$, summarized across anchors.
This measures how the Gram changes between two perturbation scales.

\subsection{Held-out directions and error definitions}
For each anchor, draw eight independent normalized Gaussian vectors
$v_j$ on the unit sphere in raw half-range coordinates, using seed
$20261021+\text{sample ID}$ (plus 100000 for Reacher). Define
\[
 u_j=\operatorname{diag}(h_1,\ldots,h_d)v_j,\qquad
 \hat r_j=\sqrt{u_j^\top G_\epsilon u_j},\qquad
 r_j=\left\|\frac{F(u_j)-F(-u_j)}{2}\right\|.
\]
The same $v_j$ are reused across scales. No held-out endpoint is used to
construct the coordinate Gram or choose its gain. The primary per-state
error is
\[
 \operatorname{WAPE}
 =\frac{\sum_{j=1}^{8}|\hat r_j-r_j|}{\sum_{j=1}^{8}r_j}.
\]
We mark a zero denominator as invalid. Each state contributes one WAPE
value, which combines its eight directions. Paired median-difference
intervals use 2,000 episode-cluster bootstrap draws with directions held
fixed. Their scope is uncertainty across sampled states and episodes;
training-seed and direction-resampling uncertainty require additional runs.

For the one-sided check, define $r_j^\pm=\|F(\pm u_j)-F(0)\|$ and compare
the same $\hat r_j$ with both values, summing errors and actual magnitudes
over all sixteen endpoints. A centered secant can cancel effects that
remain in a one-sided displacement; the two tests are not equivalent.

\subsection{Magnitude calibration and all-scale results}
The magnitude reference is $G_{\rm iso}=\operatorname{tr}(G_{\rm env})I_d/d$.
For embedding targets we also test the unnormalized model Gram and
\[
 G_{\rm model}^{\rm oracle}
 =G_{\rm model}\frac{\operatorname{tr}(G_{\rm env})}
                       {\operatorname{tr}(G_{\rm model})}.
\]
Both references use environment trace information. The oracle correction
helps quantify gain error. Deploying this correction would require
access to the environment trace. Isotropy is
defined in normalized action coordinates, whereas directions are sampled
in raw half-range coordinates, so its predicted magnitude need not be
constant across the sampled directions. Unlike $D_G$, WAPE includes gain
and the finite-scale approximation error on new directions.

\begin{table}[ht]
\floatconts{tab:response-all}
{\caption{Median embedding response WAPE (\%) at all scales. The one-sided
column uses the environment Gram. Lower is better.}}
{\small\begin{tabular}{@{}llrrrrr@{}}
\toprule
Environment & $\epsilon$ & Env.\ Gram & Isotropic & Model & Oracle gain & One-sided\\
\midrule
Reacher & 0.001 & 17.0 & 26.2 & 33.2 & 26.6 & 32.2 \\
Reacher & 0.003 & 11.3 & 22.2 & 20.6 & 18.0 & 23.9 \\
Reacher & 0.01 & 6.8 & 20.2 & 10.3 & 9.2 & 13.8 \\
Cube & 0.001 & 28.0 & 38.2 & 56.4 & 32.3 & 42.6 \\
Cube & 0.003 & 21.6 & 34.5 & 39.1 & 25.1 & 33.9 \\
Cube & 0.01 & 13.5 & 32.9 & 23.9 & 16.9 & 22.7 \\
\bottomrule
\end{tabular}}
\end{table}

At the primary scale, the model's paired median WAPE exceeds isotropy
by roughly five to seven percentage points on Reacher and fourteen to
nineteen on Cube (95\% intervals). After oracle gain correction, Reacher's
excess error narrows to about one to three points, while Cube gains an
advantage of about four to six points. The exact intervals, in WAPE
fractions, are $[0.046,0.069]$ and $[0.138,0.189]$ before correction,
and $[0.007,0.031]$ and $[-0.065,-0.036]$ afterward, respectively.
These comparisons show why response magnitude and normalized geometry
need separate evaluation.

\begin{table}[ht]
\floatconts{tab:spaces}
{\caption{Distinct environment metrics: median normalized Gram change
from $\epsilon=0.001$ to $0.01$, and primary-scale held-out centered WAPE.
WAPE is reported as a fraction (multiply by 100 for percent).}}
{\small\begin{tabular}{@{}llrr@{}}
\toprule
Environment & Output metric & Gram change & WAPE\\
\midrule
Reacher & Physical pose & $9.32\times10^{-7}$ & $2.97\times10^{-9}$ \\
Reacher & Pose + velocity & $2.11\times10^{-6}$ & $6.37\times10^{-9}$ \\
Reacher & Pixels & 0.1125 & 0.0498 \\
Reacher & Embedding & 0.4331 & 0.1702 \\
Cube & Physical pose & 0.0270 & 0.0004 \\
Cube & Pose + velocity & 0.0266 & 0.0004 \\
Cube & Pixels & 0.0808 & 0.1483 \\
Cube & Embedding & 0.4853 & 0.2801 \\
\bottomrule
\end{tabular}}
\end{table}

\paragraph{Retained failures and interpretation.}
One Reacher state has a large physical-prediction error: sample 389
reaches about 117\% WAPE at the primary scale. The other 457 states are
below 1\%. Several endpoints show a much larger angular response;
the physical cause remains unresolved. Cube's physical-pose cross-scale
Gram change has upper quartile 0.111 and maximum 1.172 despite a median
of 0.027. All such states remain in the analysis. Cube's median
physical-pose WAPE rises to about 1.47\% at $\epsilon=0.01$. These results qualify median summaries
and preclude attributing every visual scale effect solely to rendering.

\begin{figure}[ht]
\floatconts{fig:response-validation}
{\caption{Left: how much the environment Gram changes with perturbation
scale in each output space. Right: error in predicting held-out embedding
response magnitudes. Lines show medians. Reacher v2 and Cube v3 identify
replay protocols; each uses its environment's original LeWM checkpoint.
Oracle references use the environment Gram's trace.}}
{\includegraphics[draft=false,width=\linewidth]{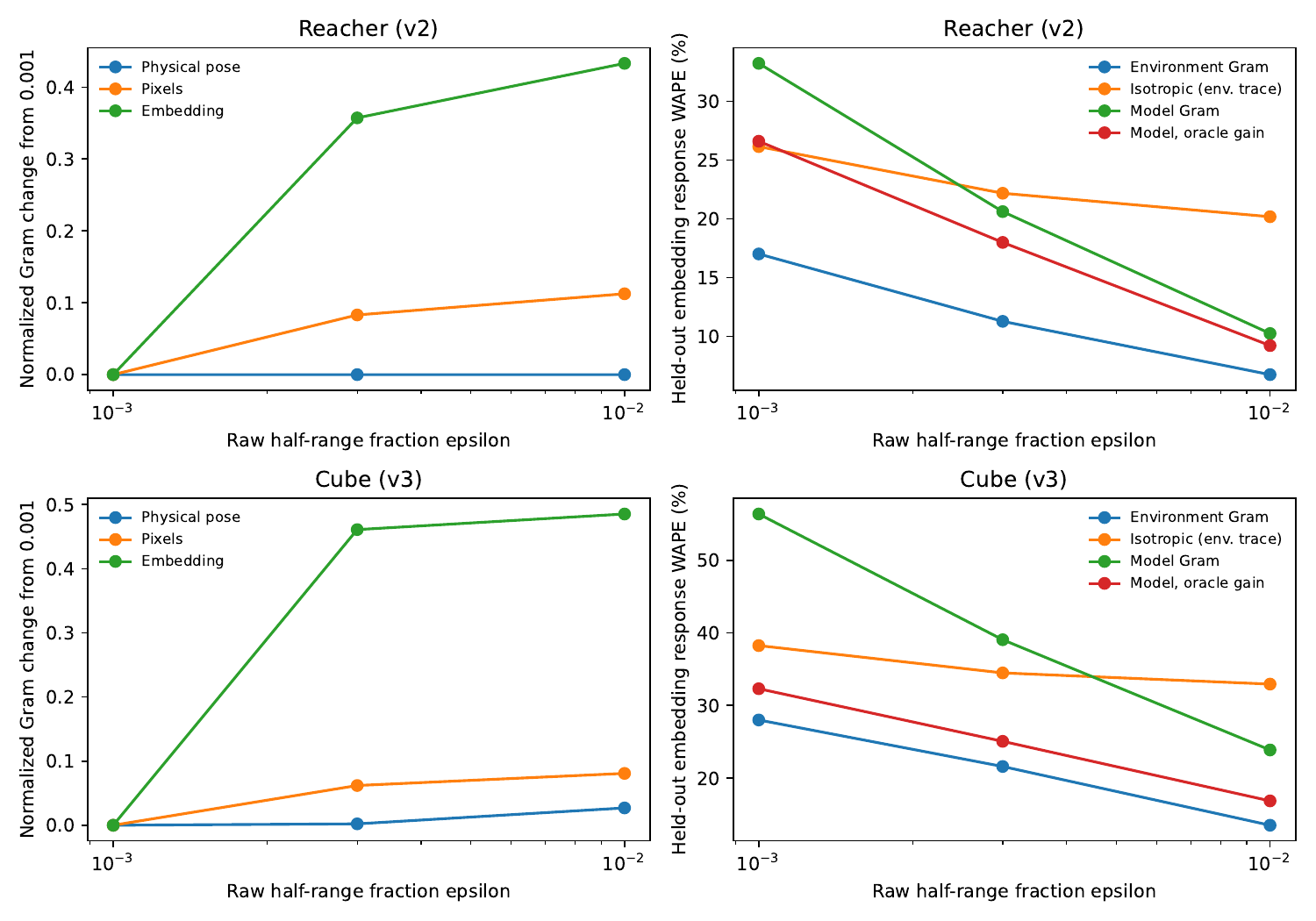}}
\end{figure}

\subsection{Replay protocol amendment and verification}
\label{apd:response-replay}
The primary discrepancy bank and its 50-anchor repeatability checks are
distinct from this subsequent response-validation run. The latter retains
all 458 Reacher and 500 Cube anchors, with 10,992 and 12,000 direction--scale
pairs. Fresh environment coordinate endpoints are evaluated alongside
held-out endpoints. Reacher uses the v2 protocol with exact parent context
and baseline images and had no replay exceptions.

Cube's initial strict replay attempts encountered intermittent one-level
RGB differences, first at a coordinate endpoint and later in context
images. Partial v1/v2 runs are archived and are not pooled with production.
The separately versioned v3 protocol permits at most one RGB intensity
level of parent context/baseline difference, records the changed values,
and recomputes model responses from the fresh context. Relative model
response change versus the parent must not exceed 1\%; baseline embedding
replay change divided by the coordinate centered-response norm must not
exceed 1\%. Coordinate image mismatches trigger repeated rollouts with
the same one-level image and 1\% embedding-response gates. Physical
qpos/qvel tolerances remain $10^{-12}$. We introduced these gates after
observing replay differences.

Production v3 stopped at anchor 292 with a baseline embedding replay ratio
of 1.193\%, above the fixed 1\% threshold. An isolated unchanged rerun
matched exactly; one unchanged production resume then completed the bank.
The thresholds and full set of anchors remained unchanged.
This retry qualifies cross-run reproducibility; the cause of the
intermittent discrepancy remains unknown.

In accepted v3 outputs, two contexts changed by at most three scalar RGB
values, each by one level; baseline images matched exactly. Maximum
relative model-response change was 0.02525\%, and maximum normalized
model-Gram change was $4.34\times10^{-5}$. Ten coordinate audit records
across six states had maximum relative embedding replay difference
0.32462\%; repeated qpos/qvel errors were zero. These accepted-output
bounds describe the completed outputs. The earlier failure remains
part of the reproducibility record. Five dedicated response-validation tests pass. Independent
analysis verified artifact hashes and reconstructed Gram predictions and
pixel/embedding response magnitudes, with maximum reconstruction error zero.


\end{document}